\documentclass{article}[10pt]

\usepackage[left=26mm,top=26mm,right=26mm,bottom=26mm]{geometry}
\usepackage[utf8]{inputenc} 
\usepackage[T1]{fontenc}    
\usepackage{lmodern}
\usepackage{hyperref}       
\usepackage{url}            
\usepackage{booktabs}       
\usepackage{amsfonts}       
\usepackage{nicefrac}       
\usepackage{microtype,xspace,bm}      
\usepackage{amsmath}
\usepackage{algorithm,algorithmic}
\usepackage{subfigure}
\usepackage{epsf}
\usepackage{psfrag}
\usepackage{color,etoolbox}
\usepackage{subfigure}
\usepackage{float}
\usepackage{multirow}
\newcommand{\mspan}{SPAN\xspace}
\usepackage{graphicx}
\graphicspath{ {img/} }

\newtheorem{defn}{Definition}[section]

\newcommand{\x}{\bm{x}}
\renewcommand{\v}{\bm{v}}

\AtBeginDocument{\setlength{\parindent}{1em}}

\begin{document}

\title{Learning Functions over Sets via Permutation Adversarial Networks}

\author{%
	Chirag Pabbaraju \\
	Microsoft Research, India \\
	\texttt{chiragramdas@gmail.com} \\
	\and
	Prateek Jain \\
	Microsoft Research, India \\
	\texttt{prajain@microsoft.com} \\
}

\maketitle

\begin{abstract}
In this paper, we consider the problem of learning functions over sets, i.e., functions that are invariant to permutations of input set items. Recent approaches of pooling individual element embeddings \cite{zaheer2017deep} can necessitate extremely large embedding sizes for challenging functions. 
We address this challenge by allowing standard neural networks like LSTMs to succinctly capture the function over the set. However, to ensure invariance with respect to permutations of set elements, we propose a novel architecture called \mspan that simultaneously learns the function as well as adversarial or worst-case permutations for each input set. The learning problem reduces to a min-max optimization problem that is solved via a simple alternating block coordinate descent technique.  We conduct extensive experiments on a variety of set-learning tasks and demonstrate that \mspan learns nearly permutation-invariant functions while still ensuring accuracy on test data. On a variety of tasks sampled from the domains of statistics, graph functions and linear algebra, we show that our method can significantly outperform state-of-the-art methods such as DeepSets \cite{zaheer2017deep} and Janossy Pooling \cite{murphy2018janossy}. Finally, we present a case study of how learning set-functions can help extract powerful features for  recommendation systems, and show that such a method can be as much as $2\%$ more accurate than carefully hand-tuned features on a real-world recommendation system.
\end{abstract}

\section{Introduction}
\label{sec:intro}
\hspace{1em}Inputs to several ML applications are naturally structured as sets. Traditionally, ML algorithms assume the inputs to be vector-valued, and thus are in general ill-suited to handle set-valued inputs. But, several recent works have shown that certain simple functions over sets can be learned accurately. 

The key insight behind these methods was discovered by \cite{zaheer2017deep, wood1996representation}, which show that {\em any} set function  can be represented as $f(X=\{\x_1, \dots, \x_n\})=\sigma(\sum_i \phi(\x_i))$, where $\phi(\x)$ is the embedding of $x$ and $\sigma$ is an arbitrary function. While this representation admits any set function, it forces to pool together independent embeddings of each set item which in turn might render the embedding layer ineffective or require it to be exponentially large for functions such as $f(X)=\max_{ij}\|\x_i-\x_j\|$. \cite{murphy2018janossy} attempts to address this challenge by casting the problem as set function learning over pairs or higher-order tuples of set elements. Naturally, for large sets, learning functions with even $3$-sized tuples becomes prohibitively expensive. 

We propose \mspan that admits arbitrary functions rather than forcing permutation-invariance explicitly which restricts the form of the function and can compromise learnability. \mspan can be trained to {\em maximize} permutation-invariance.  Naively, such a training would require optimization with respect to {\em all} the permutations of the sets in training data that would be prohibitively expensive. \cite{murphy2018janossy} proposed optimizing w.r.t. only a few random permutations. However, such an approach in general can produce large variance in the function value w.r.t. different permutations of the same set (see Section~\ref{sec:kary}). 

In contrast, our approach is to set up the problem as an adversarial network or a min-max problem, where the adversary picks up worst case permutations from a permutation network for a given function and the learner attempts to minimize the loss with respect to such permutations. That is, we set up a min-max loss function where the maximum is taken over the set of all permutations while the minimum is computed over the function. Now, several challenging set-functions can be computed by iterating sequentially over the set elements. Hence we select recurrent neural networks as the base function that is learned in a permutation invariant manner by the above mentioned min-max approach. We learn permutation functions using the Sinkhorn update based technique introduced by \cite{adams2011ranking}. 

To demonstrate an application, consider a problem where each set is generated from the spiked covariance model \cite{johnstone2001distribution}, i.e., $X=\{x_1, \dots, x_n\}$ where $\x_i \sim \mathcal{N}(0, \v\v^T + \sigma^2 I)$ and the goal is to compute $\v$, the top eigenvector of $C=\sum_i \x_i \x_i^T$. An intuitive embedding function for DeepSets \cite{zaheer2017deep} and similar approaches would be $\phi(\x_i)=\x_i \x_i^T$, i.e., the pooling layer would compute the covariance matrix $C$. However, in that case, the final layer $\sigma(\cdot)$ (typically a fixed depth feed-forward network) would be required to compute the largest eigenvector of $C$, which is challenging as it is well-known that eigenvector computation requires an iterative algorithm. On the other hand, for any permutation, our technique can apply a simple RNN transformation: ${\bm  h}_{t+1}={\bm h}_t + \v^T{\bm h}_t\cdot \v$, which is the standard Oja algorithm and is known to converge to $\v$ at nearly optimal rate \cite{jain2016streaming}. 

Independent of our work, \cite{zhang2019permoptim} proposed an algorithm that also learns permutations over the inputs, and then the permuted inputs are passed into an LSTM for predicting the label. However, the goal of our method is to learn permutation invariant functions, so we optimize w.r.t. the worst possible permutations. In contrast, \cite{zhang2019permoptim} attempts to find a permutation function which will lead to minimum loss on the training data. Hence, if the data exhibits bias w.r.t. permutations, then the method can learn that bias. E.g., consider the task of learning maximum of a given set of numbers. Suppose each set in the training data is sorted, then an identity permutation of the data combined with an LSTM that predicts the last element in the sequence would give $0$ training error. Hence, the method will be stuck at the identity permutation and learn an incorrect model. In contrast, our method will attempt to find the worst case permutation and hence would be somewhat agnostic to the training data permutations. 

We applied our algorithm to a variety of tasks including set-statistics computation, finding a tuple of $k$-farthest points, finding maximum flow in a graph etc. For each task, we compare our method against DeepSets \cite{zaheer2017deep} and other appropriate baseline algorithms. We observe that for most of the tasks, our method could significantly outperform the existing methods, especially for challenging tasks. For example, our method is $\sim$ 55\% more accurate than DeepSets for the $k$-farthest point task with $k=3$, while \cite{murphy2018janossy}'s method was difficult to scale because of a cubic number of tuples of the set. 

Finally, while set-function learning methods for computing set-statistics have been reasonably successful, they have not been demonstrated on  many large-scale real-world problems. We propose application of set-function learning methods in recommendation systems, where statistics of users' interaction with the system are represented as sets. We use  \mspan   to compute set-representations/features that can then be consumed by traditional recommendation systems. We show that such set representation learning can improve the performance of a real-world recommendation system (deployed in production) by as much as 2\% when compared to a system that uses carefully-designed features. 

\subsection{Related Work}

\hspace{1em}{\bf Applications}: A wide variety of problems can be formulated as learning functions on sets. Examples include learning a probability distribution from a set of points \cite{oliva2013distribution}, multiple instance learning \cite{maron1998framework}, 3d point classification~\cite{qi2017pointnet}, etc. In fact, several recent papers demonstrated that DeepSets \cite{zaheer2017deep} and similar pooling approaches can lead to interesting solutions for various problems in the domain of computer vision \cite{qi2017pointnet, qi2017pointnet++, Wu_2015_CVPR, shi2015deeppano} and graph convolutions \cite{kipf2016semi}, \cite{atwood2016diff}, \cite{hamilton2017inductive}. \cite{Rezatofighi_2017_ICCV} considers a complementary problem of predicting sets as outputs, by defining a likelihood over cardinalities and sets. Finally, \cite{edwards2016towards} extended Variational Autoencoders \cite{kingma2013auto} to learn summarized statistics from sets in an unsupervised manner. 

{\bf Permutation learning}: Several recent works have discussed neural-network architectures to learn permutations. In particular, \cite{adams2011ranking, mena2018learning} described how Sinkhorn normalization \cite{sinkhorn1964relationship} can be employed to learn approximate matchings in $O(n^2)$ time instead of $O(n^3)$ and applied it to ranking and ordering tasks like image ordering \cite{santa2017deeppermnet}, jigsaw puzzle solving \cite{mena2018learning}, etc. \cite{vinyals2015order} uses a different approach to permutation learning and computes "hard" assignments. Here, we note that our goal is to learn a permutation invariant set function instead of learning a specific permutation for a given input set. 

{\bf Min-max optimization}: Min-max formulations are getting increasingly popular in the ML literature with GANs \cite{goodfellow2014generative} being the most popular example. Other instances include robust learning \cite{madry2018towards} and learning with non-decomposable losses \cite{rafique2018non}. Several interesting algorithms have been proposed to solve such problems that can provide strong convergence guarantees in various settings \cite{jin2019minmax, davis2018complexity, rafique2018non}. In this work, we focus on a simple alternating block coordinate descent type of method.

\section{Problem Formulation and Method}
\hspace{1em}Suppose we are given a training dataset $\mathcal{X}=\{(X_1, y_1), \dots, (X_N, y_N)\}$ where $X_i=\{\x_1^i, \dots, \x_n^i\}$, $\x_j^i\in \mathbb{R}^d$, $y_i \in\mathbb{R}^L$, and $L$ is the dimensionality of the prediction-space.  The goal is to learn a permutation-invariant function $f: \mathbb{R}^{d\times n}\rightarrow \mathbb{R}^L$ s.t. $\sum_i \ell(y_i, f(X_i))$ is minimized where $\ell: \mathbb{R}^{L\times L}\rightarrow \mathbb{R}$ is a loss function. Permutation invariant functions are defined below: 
\begin{defn}\label{defn:perminv}
	Denote $\Pi$ as the set of all permutations, then function $f(\{\x_1, \dots, \x_n\})$ is permutation invariant if for every permutation $P \in \Pi$, the following holds: $f(X=\{\x_1, \dots, \x_n\})=f(PX)$, where $f(PX):=f(\{\x_{P(1)}, \dots, \x_{P(n)}\})$.  
\end{defn}
Most of the existing methods \cite{zaheer2017deep} explicitly enforce permutation invariance by using a pooling layer. That is, $f(X)=\sigma(POOL( \phi(\x_1,), \dots, \phi(\x_n)))$ where $POOL()$ is a simple operation like sum or max of all the elements. While this  form capture {\em all} set functions, learning with such functional restrictions can require large embedding dimension of $\phi(\cdot)$ or highly complicated function $\sigma$. 

In contrast, our method attempts to learn a permutation-invariant function without explicitly enforcing it. To this end, we first formulate the problem of learning a permutation-invariant function as the following min-max problem: 
\begin{equation}\label{eq:form1}
\min_f \max_{P_1 \in \Pi, \dots, P_N\in \Pi} \sum_i \ell(f(P_i X_i), y_i), 
\end{equation}
where $\ell(\cdot)$ is a loss function. Note that for realizable cases, i.e., when $\exists f^*$ s.t., $y_i=f^*(X_i), \forall i$, the optimal solution of \eqref{eq:form1} is guaranteed to be permutation-invariant over the {\em training} data and assuming large enough $N$, it should be nearly permutation invariant over the test data as well. Note the philosophical difference between such a formulation and that of pooling-based methods \cite{zaheer2017deep, murphy2018janossy}, which is that the later methods force permutation invariance by design and hence, even if $y_i$'s are not derived from set functions, they still lead to permutation invariant functions. In contrast, the above formulation assumes that $y_i$'s are based on permutation invariant functions. Also, this is in contrast to the formulation by an independent work by \cite{zhang2019permoptim} which attempts to learn {\em one} permutation s.t. $f(P_i X_i)=y_i$, while we require $f(P_i X_i)=y_i$ for {\em all} permutations. 

Note that solving \eqref{eq:form1} is quite challenging as it requires optimizing over the combinatorially large set of permutation matrices for {\em each} training point. Furthermore, such formulation might lead to large sample complexity. We alleviate this concern by using the technique from \cite{mena2018learning} which parameterizes permutation learning using a neural network. Using such a permutation network with \eqref{eq:form1}, we obtain the following learning problem: 
\begin{equation}
\label{eq:form2}
\min_f \max_{PN, P_i=PN(X_i)} \sum_i \ell(f(P_i X_i), y_i), \text{ s.t.},\ PN(X_i; W_{PN})=(\arg\max_{P\in \Pi} \langle P, Relu(X_i\cdot W_{PN})\rangle)^{-1},
\end{equation}
where $Relu(a)=\max(0,a)$, $W_{PN}\in \mathbb{R}^{d\times n}$. That is $PN(X; W_{PN})$ embeds $X$ in a $n\times n$ dimensional space and then learns a permutation for $X$. Finally, we parameterize $f$ as an LSTM as it can ``summarize" the set well and learn key properties about it. That is, the final optimization problem is given by: 
{\small
\begin{equation}
\label{eq:form3}
 \min_{\theta} \max_{PN, P_i=PN(X_i)} \sum_i \ell(LSTM(P_i X_i), y_i; \theta), \text{ s.t.},\ PN(X_i; W_{PN})=(\arg\max_{P\in \Pi} \langle P, Relu(X_i\cdot W_{PN})\rangle)^{-1},
\end{equation}}
where $\theta$ are the parameters of the LSTM \cite{hochreiter1997long}. That is, the formulation optimizes for LSTM's parameters $\theta$ where the data is permuted by an adversarial permutation network. 

{\bf Training Procedure}: We now discuss an optimization algorithm for \eqref{eq:form3}. We handle the min-max form using standard alternating block coordinate descent approach to reach a saddle point of the min-max problem. That is, for a fixed permutation network, we optimize for the LSTM's parameters $\theta$ to minimize the loss in \eqref{eq:form3}. To this end, we use standard back-propagation for optimizing w.r.t. $\theta$. 

Next we optimize for $PN$ for a fixed $\theta$, which requires computing gradient w.r.t. permutations which are discrete objects. Concretely, the key challenge is to compute the derivative of $M(X_i)=\arg\max_{P\in \Pi} \langle P, Relu(X_i\cdot W_{PN})\rangle$; note that $M$ is a matching of $Relu(X_i\cdot W_{PN})$. To this end, we use the method proposed by \cite{adams2011ranking, mena2018learning}  that views permutations as a point in the convex-hull of doubly-stochastic matrices that admits easy projection using iterative row and column normalization. That is, $M(X_i)=\lim_{\ell\rightarrow \inf}S(X)$ where $S^\ell(X)=\mathcal{T}_c(\mathcal{T}_r(S^{l-1}(X)))$ where $\mathcal{T}_c(\cdot)$ and $\mathcal{T}_r(\cdot)$ are the column and row normalization operators, respectively. That is, $\mathcal{T}_c(X)$ (and $\mathcal{T}_r(X)$) set the sum of each column (and row) to be $1$. Hence, $ \nabla_{W_{PN}}M(X_i)=\frac{\partial \mathcal{T}_c\odot \mathcal{T}_r}{\partial S^{l-1}(X)}\cdot \frac{\partial S^{l-1}(X)}{\partial W_{PN}}$. Now, $\frac{\partial \mathcal{T}_c\odot \mathcal{T}_r}{\partial S^{l-1}(X)}$  can be computed using a closed form expression given by \cite{adams2011ranking}. 
\begin{figure}
	
\begin{minipage}{.44\columnwidth}
		\includegraphics[width=\columnwidth]{architecture.pdf}
\end{minipage}
\begin{minipage}{.55\columnwidth}
		\begin{algorithmic}[1]
			\STATE {\bf Input}: training data $(X_i, y_i)$, inner max-iteration $K$
			\STATE {\bf Initialize} $W_{PN}^0$ using Xavier initialization \cite{glorot2010understanding}
			\FORALL{$t=1,2,\dots,$}
			\STATE {\small $P_i^T\leftarrow \arg\max_{P\in \Pi} \langle P, Relu(X_i\cdot W_{PN}^{t-1})\rangle$}
			\STATE $\theta^t\leftarrow$ SGD updates for\\  {\small $\min_\theta f(\theta)=\sum_i \ell(LSTM(P_i X_i), y_i; \theta)$ }
			\STATE $W_{PN}^t\leftarrow $ SGD+Sinkhorn updates \cite{adams2011ranking} for\\ {\small $\max_{W_{PN}} g(W_{PN})=\sum_i \ell(LSTM(P_i X_i), y_i; \theta^t)$}\\  {\small s.t.$\ \ \ P_i^T\leftarrow \arg\max_{P\in \Pi} \langle P, Relu(X_i\cdot W_{PN})\rangle$}
			\ENDFOR
		\end{algorithmic}	\vspace*{-5pt}
\end{minipage}
\caption{{\bf (a)} SPAN Architecture: It computes adversarial permutation $P^{-1}\approx P^T$ for a given point $X$ using the permutation network along with Sinkhorn updates \cite{adams2011ranking}. The data permuted by $P^T$ is then processed by the LSTM and a final feed-forward layer to predict the label $y$. {\bf (b)} Outline of the Min-max Block Coordinate Descent Method for optimizing \eqref{eq:form3}}
\label{fig:architecture}\vspace*{-10pt}
\end{figure}

{\bf Inference Procedure}: Given a test set $X$, we infer the function value as $f(X)=LSTM(P^TX, \theta)$ where $P=\arg\max _{P\in \Pi} \langle P, Relu(X_i\cdot W_{PN})\rangle$. 
As in the  training procedure, we compute  an ``incomplete" matching using a few iterations of the Sinkhorn operator (row-column wise normalization) which can be computed in $O(n^2)$ time instead of $O(n^3)$ time required by exact matching. Naturally, the solution is only  approximately correct and hence  approximating $P^{-1}$ by $P^T$ can lead to further error. However, our empirical results suggest that the approximation error is not significant as the overall method is fairly accurate. 

	\section{DeepStats: Case Study}
\hspace{1em}Consider the following recommendation problem that is critical for various social networks: compute relevance of a message $M$ written by a certain author $A_j$ posted in a particular group/channel/thread $G_k$ for a given user $U_i$. There are two critical signals in this problem: a) text of the message, b) user's affinity for an author or a group. While the message content is important, it can be captured using standard techniques like word2vec \cite{mikolov2013efficient}, doc2vec \cite{le2014distributed}. So we ignore this signal in this section. 

Now more relevant signals for this work are the indicators of a user's affinity for a group/author. This information is available in the form of a {\em set} of likes/replies/clicks by a user on messages posted by an author or on messages posted in a certain group. Hence, learning set representations is important in capturing the full strength of these signals. 
Current techniques typically compute hand-designed set-features. For example, a similar real-world user-message recommendation system deployed by Microsoft Teams uses features like weighted average number of likes or replies by a user for an author/group. Naturally, designing such heuristic set-features is challenging and might not be able to capture key discriminating representations. 

We formulate the following set representation learning problem. Say from a four-month window, we collect the following sets for each user-author pair: $S_{ij}=\{\x_1, \dots, \x_{n_{ij}}\}$, where $\x_a\in \mathbb{R}^6$ summarizes day-wise interaction between $U_i$ and $A_j$; $x_a$ is absent in the case of no interaction on that day. First coordinate of $x_a$ denotes the day of the interaction, second the day of the week, third coordinate denotes the total number of messages from $A_j$ to $U_i$, and the remaining three denote the number of likes/replies/clicks from $U_i$ for messages by $A_j$. Similarly, we define sets $Q_{ik}$ for interactions between $U_i$ and $G_k$. While we can represent this set as a matrix, the resulting matrix is extremely sparse as most users and authors have limited interactions. This leads to heavy overfitting. In contrast, set based methods can learn strong representations even  when interaction sets have a small number of elements appearing in an arbitrary order. 

We now present our {\em DeepStats} recommendation system architecture. Given $U_i$ and a message $M$ authored by $A_j$, posted in $G_k$, we compute relevance of $M$ for $U_i$ as: $R(U_i, M)=FC(f^1(S_{ij}), f^2(Q_{ik})))$ where $FC$ is a fully connected layer and $f^1, f^2$ are two set functions that embed $S_{ij}$ and $Q_{ik}$ into a vector-space. We now train this entire pipeline, i.e., FC layer and $f^1, f^2$ end-to-end using the labeled data that indicates if $U_i$ engaged/interacted with $M$. The architecture also allows for additional signals like the text of $M$ that can be directly consumed by the FC layer.

We applied the {\em DeepStats} architecture to a message recommendation system deployed in Microsoft Teams -- an enterprise social network -- and compared it with a version of the production model (minus other signals) for a random subset of users. Over four-months' interaction data, we computed $>70K$ user-author interaction sets and $>30K$ user-group interaction sets. Thereafter, we used one week's data for training the model, where any $(U_{i}, M)$ is labeled as positive if $U_{i}$ liked/replied/clicked on message $M$, and negative if there was no action. Finally, the latest one week's data is used for evaluating three models, namely Baseline, DeepStats-DeepSets and DeepStats-SPAN; DeepStats-DeepSets (DeepStats-SPAN) uses DeepSets (SPAN) to learn set features in the above described DeepStats architecture. Baseline  corresponds to a version of the production model, which uses hand-designed features like weighted averages of number of likes/replies/clicks.

\begin{figure}[t]
	\centering
	\begin{tabular}{ccc}
	\includegraphics[width=0.32\columnwidth]{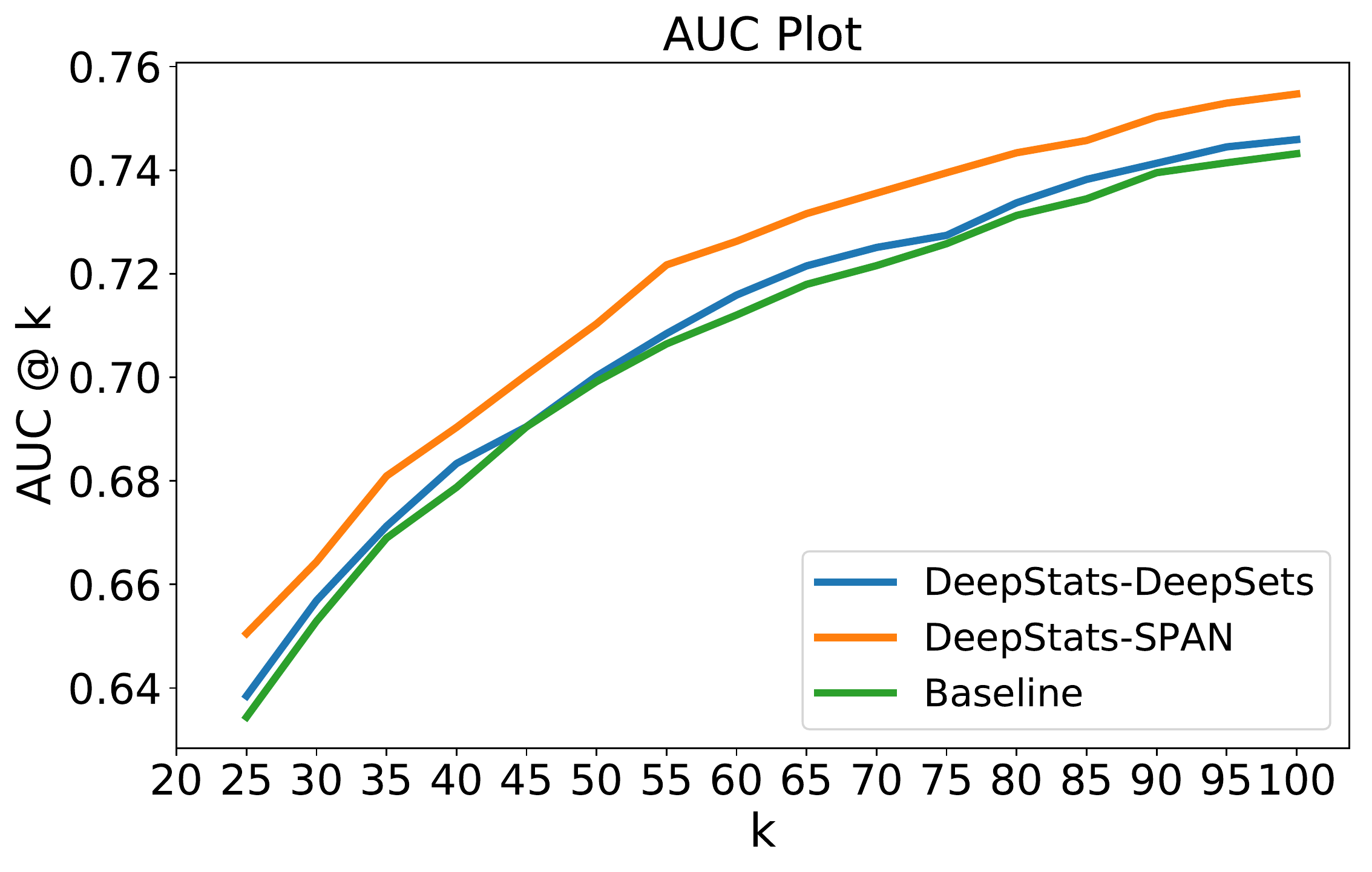}&\hspace*{-7.5pt}
	\includegraphics[width=0.32\columnwidth]{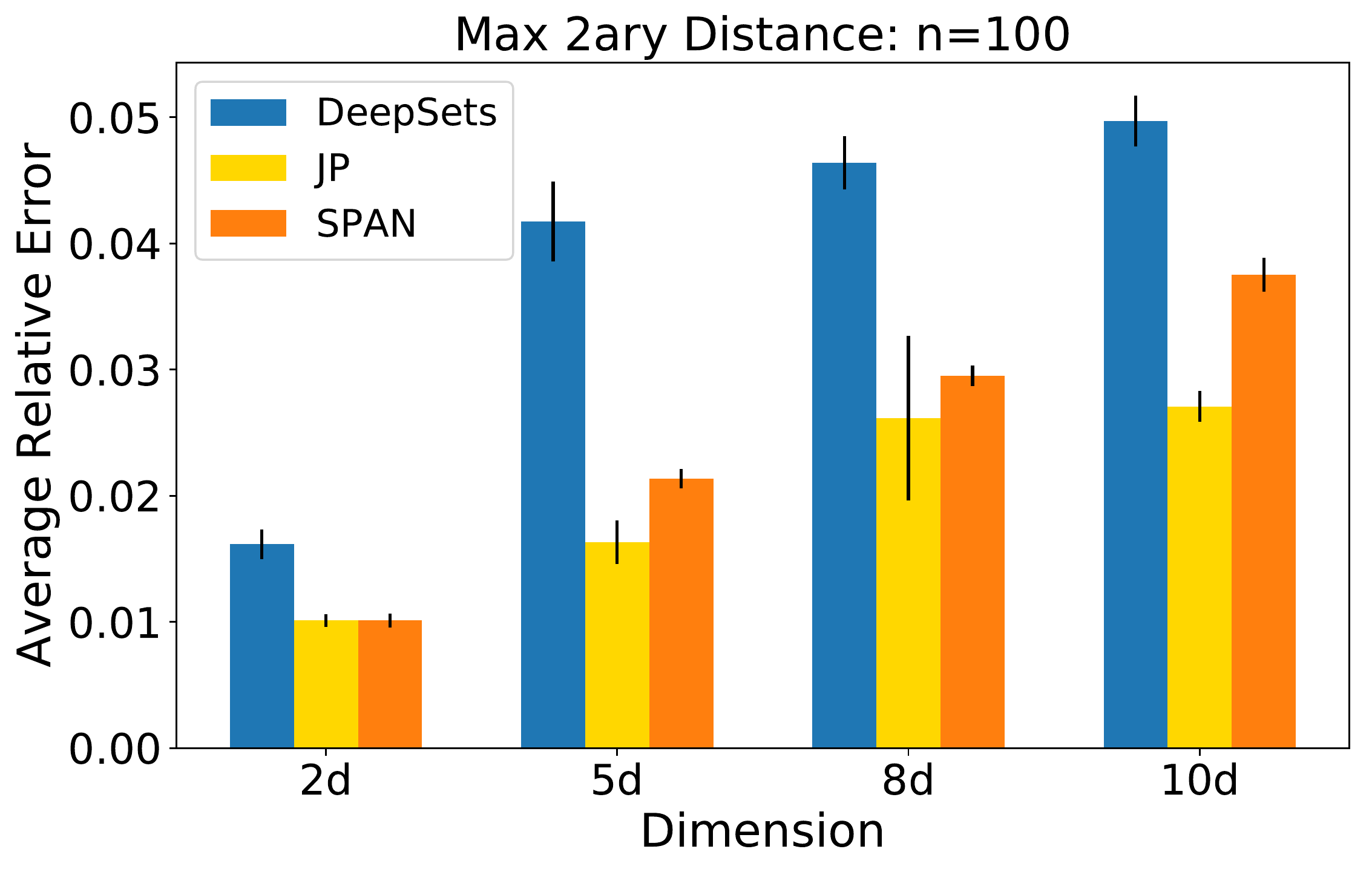}&\hspace*{-7.5pt}
	\includegraphics[width=0.32\columnwidth]{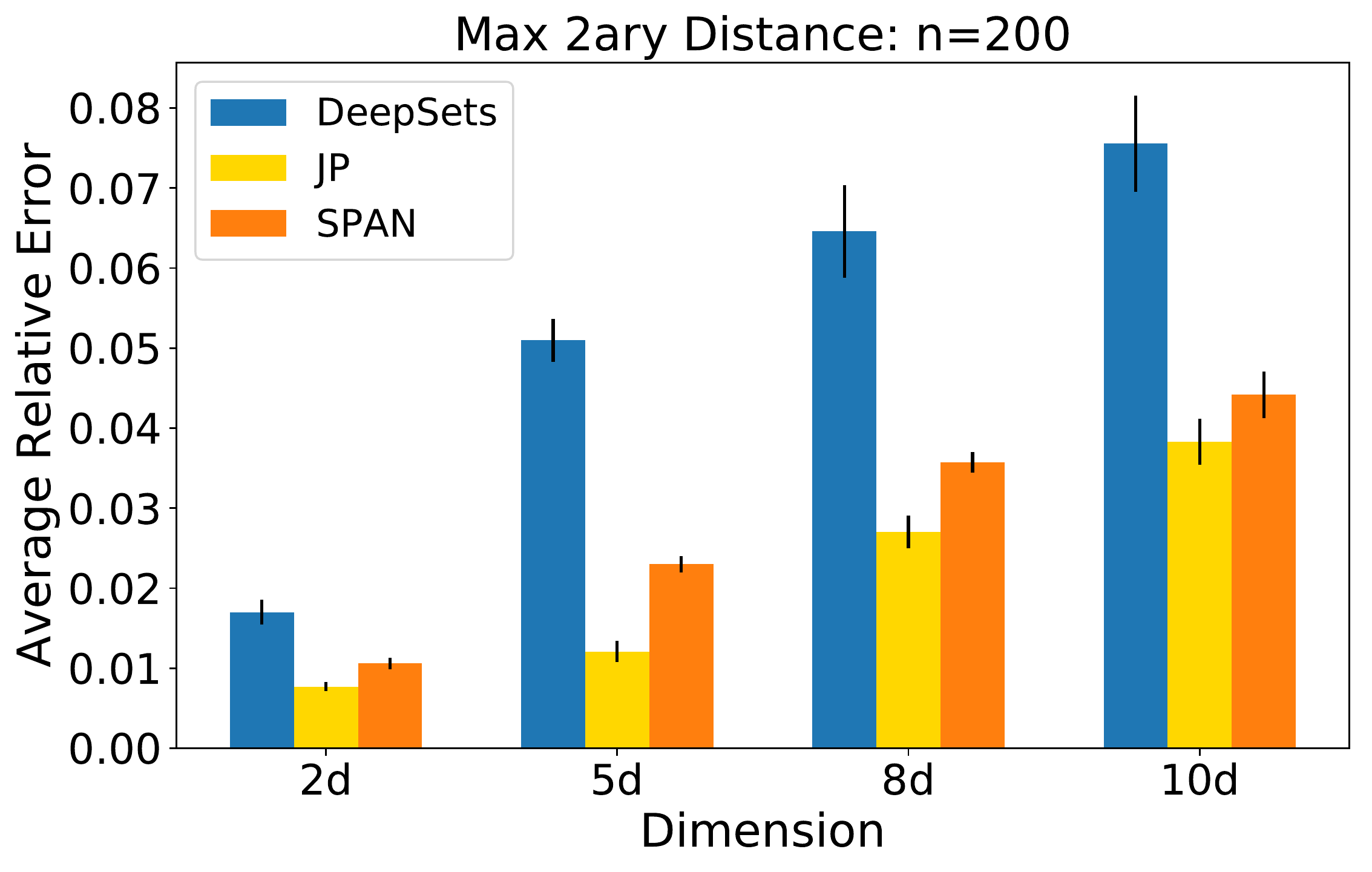}\\
	(a)&(b)&(c)\vspace*{-10pt}
	\end{tabular}
	\caption{{\bf (a)} Performance on Microsoft Teams message recommendation problem. Our DeepStats formulation with \mspan consistently achieves $\approx 2\%$ improvement over the baseline method that uses hand-designed features, in contrast, DeepStats-DeepSets improves marginally over the baseline. {\bf (b)}, {\bf (c)}: Max $2$-ary Distance: Error incurred by various methods for set sizes $n=100,200$ and different $d$ (see Section~\ref{sec:kary}). \mspan is significantly more accurate than DeepSets.}\vspace*{-15pt}
	\label{fig:teams}
\end{figure}

We compare both the methods on an AUC@k (AUC in top-$k$) metric, used by Microsoft Teams' production pipeline; relevant values of $k$ for the product are between $25$ to $100$. Figure~\ref{fig:teams} (a) shows that our proposed DeepStats-SPAN technique is able to learn a $\sim 2\%$ more accurate model than the Baseline. Furthermore, DeepStats-\mspan leads to better performance consistently compared to DeepStats-DeepSets.  

\section{Empirical Results}
\hspace{1em}In this section, we present results from experiments conducted on a variety of set-function learning tasks. In particular, the goal is a) to study performance of \mspan when compared to baseline methods on various set-function learning tasks, b) to study \mspan's performance with respect to various parameters on the tasks like number of elements in the set, dimensionality of the data, complexity of the task and c) to conduct ablation studies with respect to the two components of \mspan. 

{\bf Baselines}: We compare \mspan with DeepSets \cite{zaheer2017deep}, which is a state-of-the-art technique on general set-function learning tasks. We also compare against Janossy Pooling (JP) \cite{murphy2018janossy} which can capture $k$-ary set functions, but reduces to DeepSets when the arity is $1$ or is unknown. For the max $2$-ary distance task, we also compare against the $\pi$-SGD technique of JP \cite{murphy2018janossy}; as the performance of the method on this task was significantly worse than the other methods, we do not report it's performance on other tasks. For DeepSets, we use the standard instantiation suggested by \cite{zaheer2017deep}. That is, $f(X)=FC(\sum_i FC(x_i))$, i.e., both the embedding as well as final layers are full-connected (FC) layers. We use the same architecture for the JP method as well. 

{\bf Metric}: For each of the tasks we report absolute relative error in prediction, i.e., $\text{Err}=|y-\hat{y}|/y$ where $\hat{y}$ is the predicted value and $y$ is the ground truth. We report results averaged on $10$ random runs, and report standard deviation in the numbers as well. 

{\bf Hyperparameters}: We use a fixed set of parameters for \mspan irrespective of tasks. That is, we set the learning rate to be $1e-4$, hidden size of LSTM to be $128$, batch-size to be $32$ and FC layers to be of size 128. The temperature parameter as described in \cite{mena2018learning} is fixed to be $0.1$, and we set the number of iterations of the Sinkhorn operator on the matrix output by the permutation network to be 100. To ensure fair evaluation of DeepSets and JP, we did extensive sweeps for both the methods. In particular, we computed validation set loss over every combination of the following parameters and report numbers corresponding to parameters with the lowest loss on the validation set: a) size of the embedding and final FC layer with values in $\{64, 128\}$, b) dropout rate in each layer selected from $\{0.5, 0.2, 0.0\}$ and c) $l_2$ weight decay over weights with regularization constant in $\{0, 0.1, 0.01, 1\}$. In addition, after extensive experimentation, we found that a learning rate of $1e-4$ with the Adam optimizer \cite{kingma2014adam} seemed to consistently provide the best results. 

We now discuss various set-function learning tasks, our experimental methodology, and report results for \mspan and baseline methods. A preliminary version of the source code for SPAN is available at \url{https://github.com/chogba/SPAN}.
\subsection{Max $k$-ary Distance}\label{sec:kary}
\hspace{1em}The objective of this task is as follows: Given a set of $n$ vectors in $R^{d}$, learn that the underlying function is $f(X)= \max_{H\in [n], |H|=k}\,\sum_{(i,j)\in H}\|x_{i} - x_{j}\|$. Specifically, this task requires capturing $k$-wise relations between elements of the set. DeepSets' pooling operation discards information about relations between elements, thus leaving most of the heavy-lifting to the final layer. 

Training data for this task was generated as follows: for $k$-ary task, for {\em each set} $X$, we generate $k$ cluster centers $\mu_1, \dots, \mu_k \in \mathbb{R}^d$ where $\mu_q, 1\leq q\leq k$ are sampled from the uniform distribution i.e. $\mu_q\sim Unif[1:n]^d$. We then sample each $\x_i, 1\leq i\leq n$ from the standard mixture of Gaussians distribution, i.e., $\x_i \sim \frac{1}{k}\sum_q \mathcal{N}(\mu_q, 10\cdot I)$. We set the label as $y=\max_{H\in [n], |H|=k} \sum_{(i,j)\in H}\|x_{i} - x_{j}\|$. We studied the task for $k=2,3$. 

{\bf Max 2-ary Distance}: We generated $N=5000$ sets for training and $1000$ sets for testing with each set containing $n=100,200$ points. We conducted experiments for $d=2,5,8,10$. As mentioned above, we use a fixed set of hyperparameters for \mspan while we conduct a thorough grid search to select the best parameters for DeepSets and JP. Figure~\ref{fig:teams} (b), (c) reports the average relative error incurred by each method  averaged over the sets in test data and $10$ independent runs with  $n=100,200$, respectively. \mspan incurs significantly lower error than DeepSets. For example, for $d=5$, the error incurred by DeepSets is nearly twice the error incurred by \mspan for $n=100$. The ratio of error jumps to $2.2$ for $n=200$ which also reflects the general trend that \mspan's error gain over DeepSets increases for larger $n$, and hence for more challenging tasks. 

When compared to JP, our method incurs more error in this task. However, note that the JP 2-ary model enumerates all pairs of vectors in a set, and thereafter applies DeepSets to the significantly easier task of computing maximum over $O(n^2)$ numbers, so it explicitly captures the fact that the function is $2$-ary. However with larger $n$, the number of enumerations explodes, making it difficult to train the JP model. This aspect is explored further in the following section. We also evaluated the $\pi$-SGD variant of JP\cite{murphy2018janossy} for the case with $d=2$ by sampling and training with 20 random permutations of each set, but the results were significantly worse: for $n=100$, there was an 18x increase in average relative error, while there was a 25x increase in the case of $n=200$. Also, the predictions showed a large standard deviation $(\approx 2)$ for different permutations of the same set. Since the error numbers were larger by an order of magnitude, we do not focus on that method in the remaining sections. 
		
\begin{figure}
	\centering
	\begin{tabular}{ccc}\hspace*{-5pt}
	\includegraphics[width=.25\columnwidth]{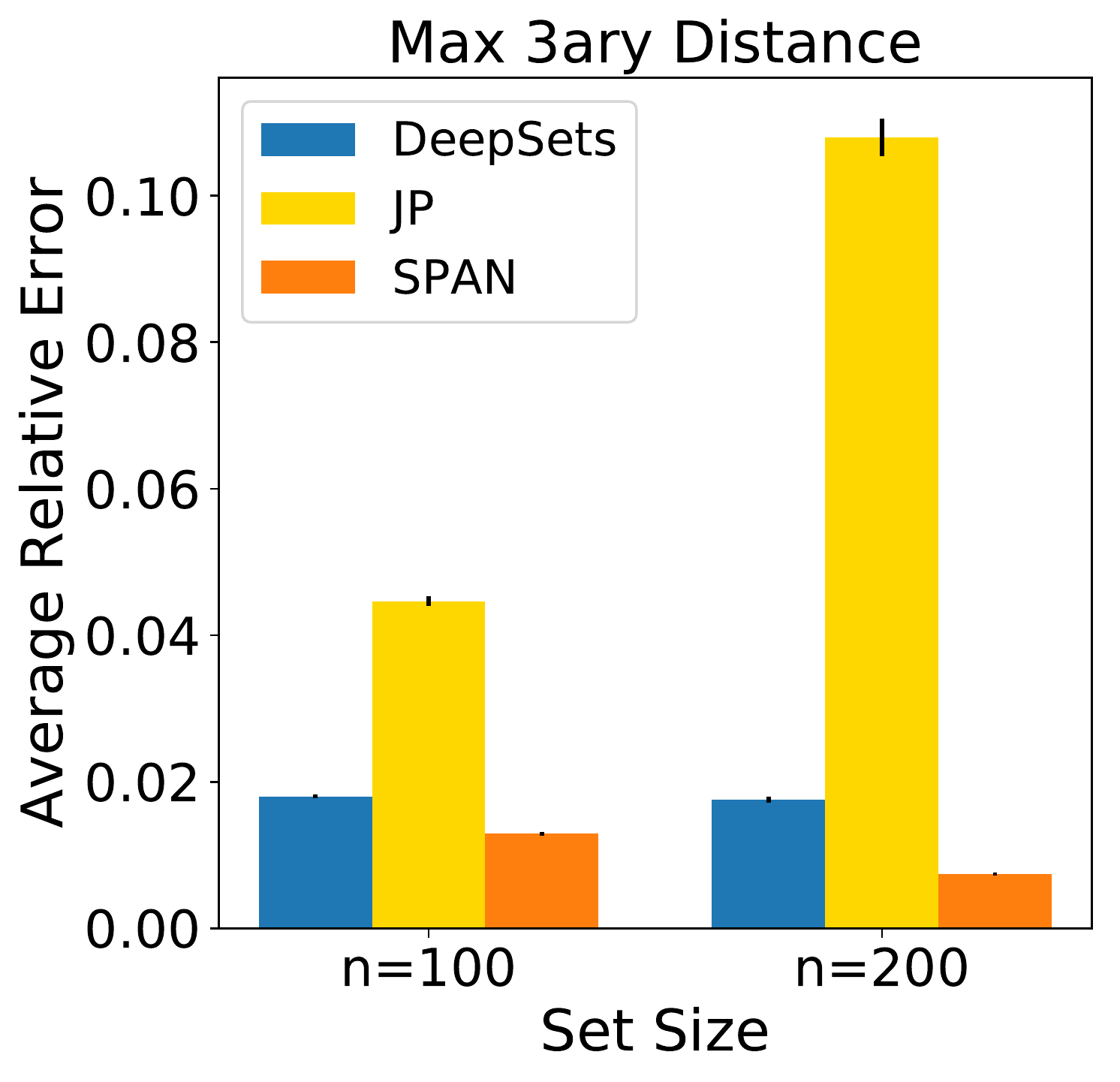}&\hspace*{-10pt}
	\includegraphics[width=.3\columnwidth]{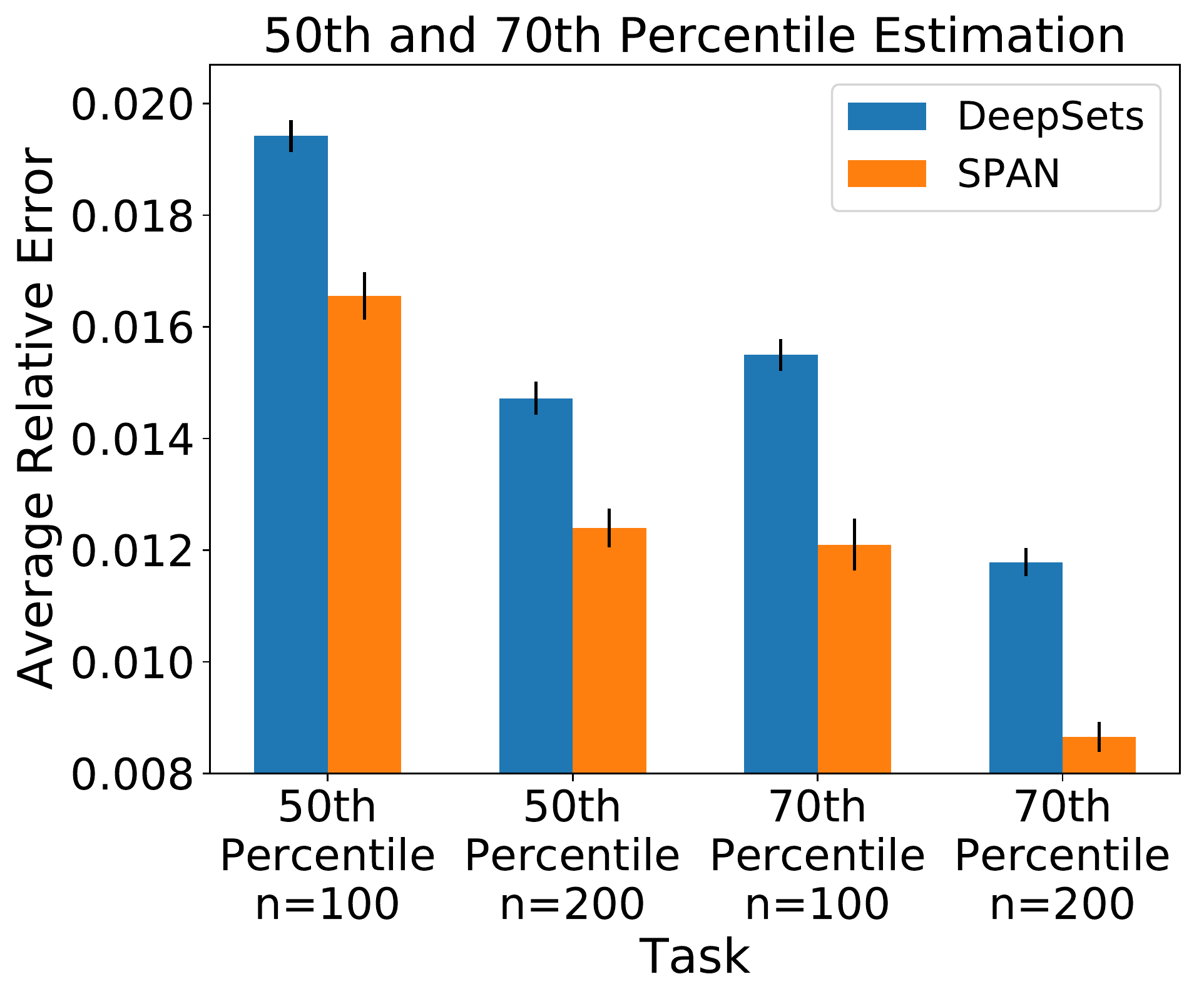}&\hspace*{-10pt}
	\raisebox{50pt}{\begin{tabular}{|c|p{40pt}|p{20pt}|}
		\hline
		& {\tiny \mspan w/o APN} & {\tiny \mspan}  \\
		\hline
		\includegraphics[scale=0.18]{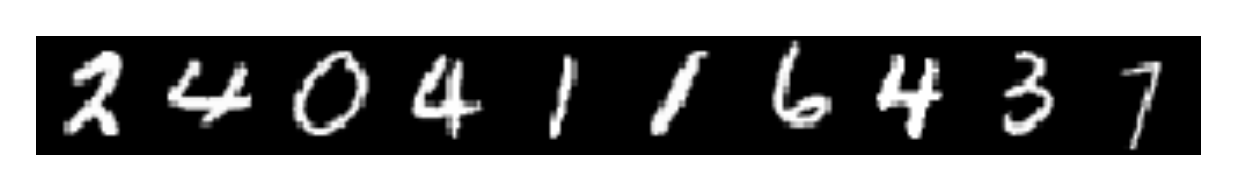} & 7 & 7 \\[-4pt]
		\includegraphics[scale=0.18]{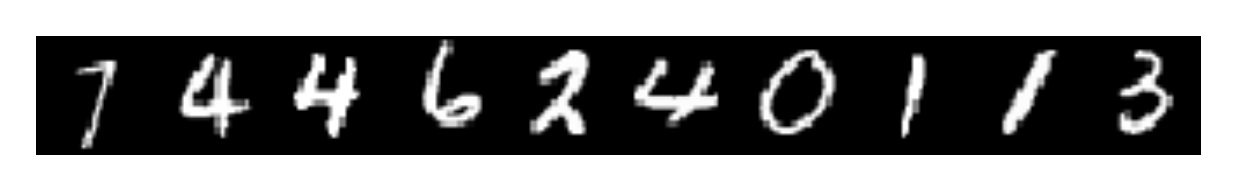} & 3 & 7 \\
		\hline
		\includegraphics[scale=0.18]{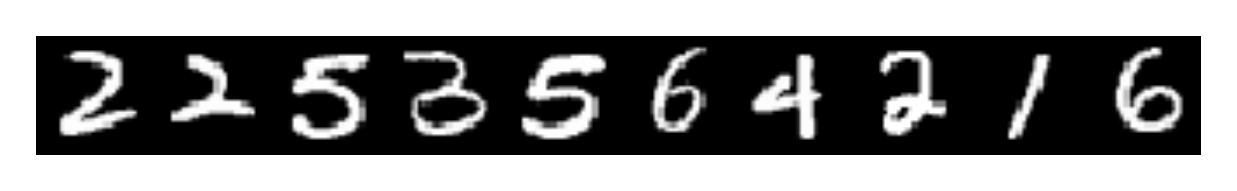} & 6 & 6 \\[-4pt]
		\includegraphics[scale=0.18]{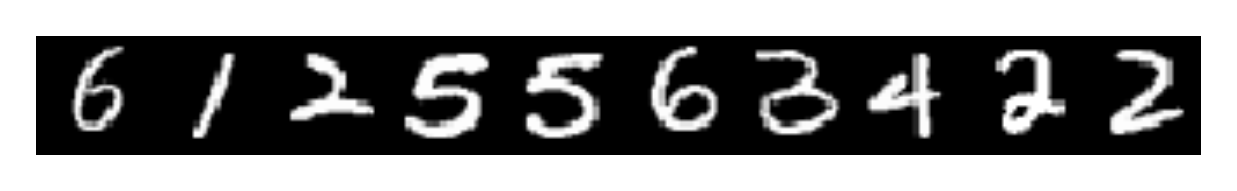} & 2 & 6 \\[0pt]
		\hline
		\includegraphics[scale=0.18]{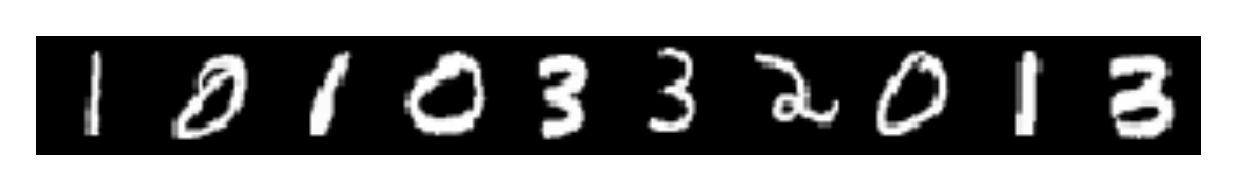} & 3 & 3 \\[-4pt]
		\includegraphics[scale=0.18]{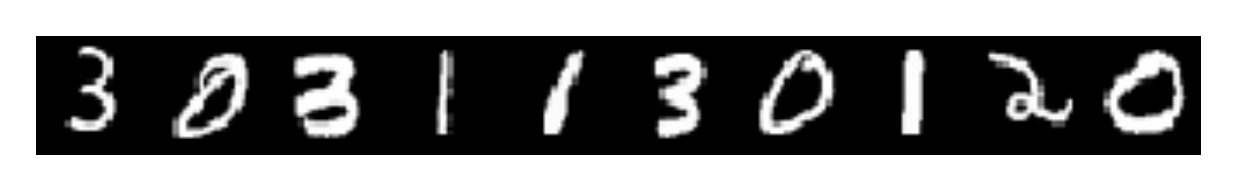} & 0 & 3 \\[0pt]
		\hline		
	\end{tabular}}\vspace*{-5pt}\\
	(a)&(b)&(c)\vspace*{-10pt}
	\end{tabular}
	\caption{{\bf (a)} Max $3$-ary Distance: Error incurred by various methods for set sizes $n=100,200$ and with $d=2$ (see Section~\ref{sec:kary}). {\bf (b)} $r^{th}$ Percentile Estimation: Relative error for various methods with set size $n=100,200$. 
	(c) Effect of Adversarial Permutation Network (APN) in \mspan: The goal here is to predict the maximum digit in a given set of MNIST images. \mspan exhibits permutation-invariance and correctly predicts the largest digit in the set of images while \mspan w/o APN gets biased to predict the last image in the set (Section~\ref{sec:ablation}).}
	\label{fig:plots}\vspace*{-5pt}
\end{figure}

		
{\bf Max 3-ary Distance}: We generate $10000$ training and $2000$ test sets as above. As even generating data and labels itself is expensive, we conduct the experiment with $d=2$ only. Figure~\ref{fig:plots} (a) compares averaged relative error for various methods. Similar to the $2$-ary task, here also \mspan is significantly more accurate than DeepSets. In fact, for $n=200$, the ratio of DeepSets' error to \mspan's error increases to $\sim 2.4$. However, results for JP are more interesting. Recall that due to $3$-arity of the function, this method will form training sets of size $\binom{n}{3} = 1313400$ (when $n=200$). Naturally, even a simple function like max with such large sets is difficult and hence the relative error for this method turns out to be an order of magnitude larger than that of \mspan. Furthermore, for $n=200$, we had to train the $3$-ary JP model for $3$ days with a batch size of just 4 to fit in the GPU memory, thus underlining challenges in capturing higher order interactions between set elements via JP method.

		
\vspace*{-5pt}
\subsection{\bf{$r^{th}$} Percentile Estimation}\label{sec:perc}\vspace*{-5pt}
\hspace{1em}In this task, the objective is to estimate the $r^{th}$ percentile value in a set of $n$ numbers. We generate random sets of integers of size $n = 100, 200$, with values ranging from $[1,100]$ for $n=100$ and $[1,200]$ for $n=200$. The label for each set was the $r^{th}$-percentile value; we consider $r=50, 70$. We train with 5000 sets, and evaluate the performance of \mspan and DeepSets on 1000 sets. Since intuitively this is a $1$-ary problem, JP reduces to DeepSets. Here again, average relative error for DeepSets is significantly higher than that of \mspan, and the difference is more pronounced for larger set size $n=200$. In particular, for $n=200$, DeepSets incurs $\sim 18\%$ more error in the $50$-th percentile learning task,  while for the $70$-th percentile learning task, DeepSet's relative error is $\sim 37\%$ higher than \mspan.

		\vspace*{-5pt}
\subsection{Multiple-source Maximum Flow Computation}\label{sec:flow}\vspace*{-5pt}
\hspace{1em}We choose the next task from the domain of submodular functions over graphs. The goal here is to learn the following set function: given a directed graph $G$, with vertex set $V$ and edge set $E$, and a capacity $c_{i}$ defined $\forall e_{i} \in E$, the function value is the maximum flow from a subset of vertices $H \subset V$ to a destination sink $s$. For this task, we generated a random directed graph with one connected component, 100 vertices, 300 edges and capacities sampled randomly in $[1,20]$. The sink $s$ was fixed to be an arbitrary vertex. We sampled $N=5000$ random subsets of 3 vertices $H_{i} \subset V \setminus s$ from the graph. For each subset $H_{i}$, we set the label to be the maximum flow $f_{i}$ from $H_i$ to $s$ computed using Ford-Fulkerson's algorithm \cite{ford2009maximal}. The labels ranged from 17 to 64. We did not scale to subsets of larger size in this task, because the distribution of labels for large subsets converges to a small range of values. The training sets $X_i$ consist of one-hot encodings of each vertex in $H_i$ concatenated with an embedding of the adjacency matrix of $G$. Naturally, the task is quite challenging, but \mspan is able to learn the task reasonably well with an average relative error of $0.015$; DeepSets' error is $30\%$ higher and small standard deviation between various runs indicates that the improvement is statistically significant (see Table~\ref{table:vtask}(a)). 
\begin{table}[t!]
\centering
\begin{tabular}{cc}
	{\small 
	\begin{tabular}{|c|c|}
	\hline
	& Average Relative Error (Std)\\
	\hline
	DeepSets & 0.0203 (0.0017)  \\
	SPAN & 0.0154 (0.0009) \\
	\hline
	\end{tabular}}&
	{\small 
	\begin{tabular}{|c|c|c|}
	\hline
	& $d=20$ (Std) & $d=30$ (Std) \\
	\hline
	DeepSets & 0.9072 (0.0032) & 0.8331 (0.0038) \\
	SPAN & 0.9338 (0.0021) & 0.8943 (0.0056) \\
	\hline
	\end{tabular}}\\
	(a)&(b)
\end{tabular}
\caption{{\bf (a)} Error incurred for multi-source max-flow task (Section~\ref{sec:flow}) with $n=3$, $|V|=100$, $|E|=300$. {\bf (b)}: Absolute cosine similarity of the predicted vector and the top eigenvector (Section~\ref{sec:spiked}).}
\label{table:vtask}\vspace*{-20pt}
\end{table}
\subsection{Learning Top-eigenvector in Spiked Covariance Model}\label{sec:spiked}
\hspace{1em}Our final task requires learning a vector-valued set function. That is, to learn eigenvectors from a given set of points sampled from the spiked covariance model \cite{johnstone2001distribution}. For each set $X_i$ we sample ${\bm v}$ uniformly at random from the unit-sphere, and each $\x_j\sim N(0, {\bm v}{\bm v}^T+\sigma^2 I)$ for all $1\leq j\leq n$ and $n=100$. Hence, ${\bm v}$ -- the top-eigenvector of the covariance matrix -- is the label for the task. The training data comprised of $10000$ such sets and the test data had $2000$ sets. For a normalized prediction vector $\widehat{\bm {v}}$ we report the absolute cosine similarity, i.e., $|{\bm v}\cdot \widehat{\bm {v}}|/(\|{\bm v}\|\cdot\|\widehat{\bm {v}}\|)$. Table~\ref{table:vtask} (b) shows that \mspan's cosine similarity is  about $3\%$ higher than DeepSets for $d=20$ and the gap increases to $6\%$ for the more challenging task of $d=30$. 

Finally, we observe that over $20$ random permutations of each set in test data for {\em each} of the above described tasks, predictions of \mspan do not change significantly, with the ratio of standard deviation in predictions to the mean being $\leq 1e-5$ (for \ref{sec:spiked} too, the predicted vectors across different permutations agree upto $1e-5$ precision in each component), thus indicating that \mspan can learn effectively permutation-invariant functions.


\vspace*{-5pt}
\subsection{Ablation Studies}\label{sec:ablation}\vspace*{-5pt}
\hspace{1em}Finally, we study the isolated impact of the two components in our architecture: adversarial permutation network (APN) and the final learner. 

{\bf APN Ablation}: Here, we remove the adversarial PN from our architecture (Figure~\ref{fig:architecture}, \eqref{eq:form3}), which then simplifies to applying a standard LSTM to the set treating it as a sequential point. We then compare performance of \mspan against \mspan w/o APN on the task of finding the maximum digit contained in a set of MNIST \cite{mnist} images. To demonstrate the impact of APN, we train both the models with biased training data where the image corresponding to the maximum digit is placed as the last element of the set; however, the test data does not exhibit this bias. The intuition behind this bias is that the LSTM, without APN, would learn to predict the last element in each set as the label instead of the max element\footnote{As mentioned in the introduction, a similar bias holds for the method by \cite{zhang2019permoptim} when initialized with the identity permutation.}. Figure~\ref{fig:plots} (c) confirms this intuition where \mspan w/o APN always picks up the last digit in the set irrespective of the maximum digit, while \mspan with APN is able to identify the maximum digit accurately. Table~\ref{table:ablationLSTMvsFC} (a) further confirms the same hypothesis as \mspan w/o APN is able to pick up the maximum digit in only 10\% of the test points. 

{\bf Learner Ablation}: Here, we study the impact of LSTM on \mspan's performance. Intuitively, LSTMs and similar recurrent networks are well-suited for set learning tasks as they allow for a compact ``summarization" of the set iteratively that can be updated after processing each point in the set. E.g., an LSTM's internal state can maintain the maximum number out of elements seen till a step, and update the state only if the next element is larger than the current state. Table~\ref{table:ablationLSTMvsFC} (b) indicates a similar trend where \mspan with LSTM's performance is better than \mspan with fully-connected (FC) layer's performance on the challenging max $3$-ary distance task (Section~\ref{sec:kary}).

\begin{table}[t!]
	\centering
	\begin{tabular}{cc}
			{\small 
			\begin{tabular}{|c|c|c|c|}
				\hline
				& Max & Last & Other \\
				\hline
				SPAN w/o APN & 0.100 & 0.576 & 0.324 \\
				SPAN  & 0.894 & 0.009 & 0.097 \\
				\hline
			\end{tabular}
		}&
		{\small 
	\begin{tabular}{|c|c|c|}
		\hline
		\multirow{2}{*}{} & \multicolumn{2}{|c|}{Average Relative Error (Std)}  \\
		\cline{2-3}
		& $n=100$ & $n=200$ \\
		\hline
		SPAN & 0.0129 (0.0002) & 0.0074 (0.0002)\\
		SPAN w/ FC & 0.0135 (0.0002) & 0.0126 (0.0005) \\
		\hline
	\end{tabular}
	}
	\\
	(a)&(b)
	\end{tabular}
	\caption{Ablation studies. {\bf (a)}: \mspan w/o APN is biased by training data in predicting the last element in the set instead of the maximum element. The numbers indicate the fraction of sets for which the model predicted the max, last and other set elements. (Section~\ref{sec:ablation}). {\bf (b)}: Comparison of SPAN with LSTM and SPAN with FC on max $3$-ary distance task with $d=2$ and $n=100, 200$.}\vspace*{-10pt}
	\label{table:ablationLSTMvsFC}
\end{table}

\section{Conclusions}
\hspace{1em}We studied the problem of learning functions over sets. We modelled the problem as a permutation adversarial network, where the goal is to find a learner (LSTM) that correctly predicts the label of a given set despite adversarial permutations of the set elements. We demonstrate that our approach is able to learn  challenging set functions like maximum flow over a graph, and can outperform existing state-of-the-art techniques like DeepSets \cite{zaheer2017deep}, Janossy Pooling \cite{murphy2018janossy}. We also present an application of our method in the domain of recommendation systems via a DeepStats architecture. Further exploration of such recommendation and similar problems is of great interest. Also, establishing sample complexity of learning permutation-invariant functions is an important open question. 

\clearpage
\bibliographystyle{plain}
\bibliography{references}

\end{document}